\documentclass[runningheads]{llncs}

\usepackage[utf8]{inputenc}
\usepackage{makeidx}  
\usepackage{graphicx}
\usepackage{tabularx}
\usepackage{graphicx}
\usepackage{booktabs}
\usepackage{url}
\usepackage{array}
\usepackage{multirow}
\usepackage{amsmath, amssymb}
\usepackage{booktabs}
\usepackage[style=lncs]{biblatex}
\usepackage[misc]{ifsym}
\bibliography{paper1169}

\usepackage{caption} 
\title{Factorised spatial representation learning: application in semi-supervised myocardial segmentation}

\author{Agisilaos Chartsias\textsuperscript{1\Letter} \and Thomas Joyce\textsuperscript{1} \and Giorgos Papanastasiou\textsuperscript{2} \and Scott Semple\textsuperscript{2} \and Michelle Williams\textsuperscript{2} \and David Newby\textsuperscript{2} \and Rohan Dharmakumar\textsuperscript{3} \and Sotirios A. Tsaftaris\textsuperscript{1}}

\institute{\textsuperscript{1}Institute for Digital Communications, School of Engineering, University of Edinburgh, West Mains Rd, Edinburgh EH9 3FB, UK.
\textsuperscript{2}Edinburgh Imaging Facility QMRI, Centre for Cardiovascular Science, Edinburgh, EH16 4TJ, UK.
\textsuperscript{3}Cedars Sinai Medical Center Los Angeles CA, USA.\\
\email{agis.chartsias@ed.ac.uk} }

\authorrunning{A. Chartsias et al.}
\titlerunning{Factorised spatial representation learning}

\begin{document}

\maketitle

\begin{abstract}

The success and generalisation of deep learning algorithms heavily depend on learning good feature representations. In medical imaging this entails representing anatomical information, as well as properties related to the specific imaging setting. Anatomical information is required to perform further analysis, whereas imaging information is key to disentangle scanner variability and potential artefacts. The ability to factorise these would allow for training algorithms only on the relevant information according to the task. To date, such factorisation has not been attempted. In this paper, we propose a methodology of latent space factorisation relying on the cycle-consistency principle. As an example application, we consider cardiac MR segmentation, where we separate information related to the myocardium from other features related to imaging and surrounding substructures. We demonstrate the proposed method's utility in a semi-supervised setting: we use very few labelled images together with many unlabelled images to train a myocardium segmentation neural network. Specifically, we achieve comparable performance to fully supervised networks using a fraction of labelled images in experiments on ACDC and a dataset from Edinburgh Imaging Facility QMRI. Code will be made available at \url{https://github.com/agis85/spatial_factorisation}.

\end{abstract}

\section{Introduction}

The effectiveness of any (deep or shallow) learning algorithm lies in learning good feature representations. These must be maximally informative for the task at hand, whilst being invariant to unrelated information (e.g. variations in imaging, noise, etc), so that they can generalise to unseen examples \cite{bengio2013representation}. 
Invariance to some factors, e.g. translations, can be attributed to the architecture, for instance with the use of convolution and max-pooling, but invariance to more complex factors is achieved by the learning process, and specifically encouraged by regularisers (explicit regularisation) or data augmentation (implicit regularisation) \cite{achille2017emergence}.

At a high level the aim is to keep relevant but discard irrelevant information, however which information is relevant is strongly task dependent. In this paper we are interested in the related task of decomposing the input into meaningful components (or factors), which
offers many benefits. Critically it enables preserving factors not directly relevant to the primary task, which may otherwise be discarded when driven by pure supervised learning. It is then possible to reuse parts of a factorised representation for related tasks or for transfer learning to other domains. Further, by capturing specific properties of the data, such representations become easier to interpret, an aspect of currently heated debate in deep learning with dedicated workshops on the topic (e.g. \url{http://interpretable.ml}). Finally, finding (and preserving) the factors of variation is \emph{de facto} necessary for generative models, in order to be able to (re)produce realistic results. 

Factorised representations are a recent topic in deep learning \cite{chen2016infogan, cheung2014discovering, higgins2016beta, kim2018disentangling, mathieu2016disentangling, siddharth2017learning}. These works focus on decomposing feature representations into discrete or continuous latent vectors.  At present, there has not been any work on learning factorised representations that include spatial components, which are of particular interest for spatially equivariant tasks using fully convolutional networks (such as segmentation and registration).\footnote{Concurrent work with ours, introduced auxiliary variables and combined them with a spatial representation for the task of image translation \cite{almahairi2018augmented, huang2018munit}.}
Here we propose a spatial decomposition network (SDNet), that decomposes input images into a spatial map containing  anatomical information and a latent vector of image intensity information (and residual anatomical information), leveraging the cycle-consistency loss~\cite{zhu2017unpaired}, originally proposed for style transfer. 
Specifically, we train two networks: one that learns a decomposition into spatial and non-spatial latent factors, and one that learns to reconstruct the input image using the decomposed representation.
We demonstrate our method in semi-supervised myocardium segmentation, using a small amount of labelled but a large pool of unlabelled cardiac cine MR images.
In this application, our method learns to decompose the shape and location of the myocardium from information related to surrounding structures and pixel intensities (related to scanner properties and other imaging characteristics).
%
%

In summary, our contributions are the following: (a) We propose a new method for disentangling images into a spatial map and a continuous vector, which is directly applicable to medical images for representing anatomical and non-anatomical information.
(b) We show properties of the decomposed latent space by generating examples using latent space arithmetic.
(c) We demonstrate the utility of our method in a semi-supervised myocardium segmentation task, where the learned high-level topological knowledge allows the network to retain performance in a low data regime.

\begin{figure*}[t]
\centering
\includegraphics[width=\textwidth]{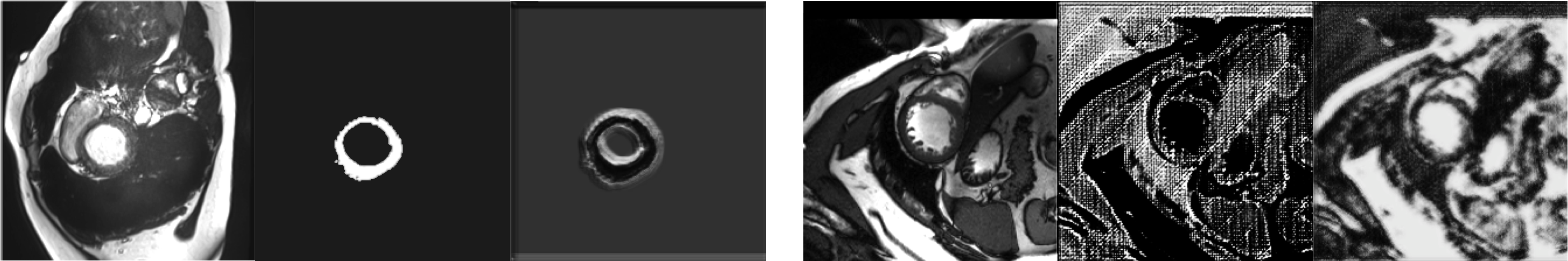}
\caption{Input images, segmentation masks and reconstructions produced by a CycleGAN. Left: high weight on segmentation, right: high weight on reconstruction.\label{fig:cyclegan_example}}
\end{figure*}

\section{Related Work}
\noindent \textbf{Learning factorised representations:} To date interest has centred on representing factors of variation as independent latent variables, using Autoencoders \cite{cheung2014discovering} or Variational Autoencoders (VAE) \cite{siddharth2017learning} to decompose classification related factors from remaining image reconstruction factors. VAE were used for unsupervised learning of factorised representations, where the factors of variation are discovered throughout the learning process \cite{higgins2016beta, kim2018disentangling}. A generative model combining VAE with Generative Adversarial Networks (GAN) was proposed in \cite{mathieu2016disentangling} to decompose the input into image classes and remaining factors. Further, InfoGAN was proposed in \cite{chen2016infogan}, in which mutual information between a latent variable and the generated images is maximised. More recently, feature decompositions were proposed for video data to separate foreground from background \cite{vondrick2016generating}, and motion from content \cite{tulyakov2017mocogan}.
These methods learn decomposed representations in terms of continuous or discrete variables; however, spatial information could be directly represented in a convolutional map, and this would be useful when the learning task is semantic segmentation. Our proposed method produces a decomposition as a combination of spatial and non-spatial information. This makes our learned representation directly applicable to  segmentation tasks. 

\noindent \textbf{Semi-supervised segmentation:} Using unlabelled data to guide learning is appealing and has been exploited by the community.  In \cite{bai2017semi} an iterative method was proposed, where a CNN is alternately trained on labelled and post-processed unlabelled sets. GANs were used in \cite{zhang2017deep}, for a gland segmentation task, involving supervised and unsupervised adversarial costs. Another approach \cite{baur2017semi} aims to minimise the distance between embeddings of labelled and unlabelled examples by comparing them in feature space.
Semi-supervised learning with GANs was also proposed for semantic segmentation. The discriminator classifies between real and synthetic segmentation masks produced by the generator in \cite{luc2016semantic}, while in \cite{souly2017semi} the generator is used to increase the dataset size and the discriminator performs segmentation. Our method differs from these in that we introduce both adversarial and cycle losses to push mask generation to be spatially aligned with the image and avoid the need for post-processing as in \cite{bai2017semi}.
Also we do not require any pairs of image and masks for discriminator training as in \cite{zhang2017deep}, and we retain all information, in contrast to \cite{baur2017semi} which preserves only task relevant information.

\section{Proposed Approach: the SDNet} \label{sec:approach}
\noindent \textbf{Motivation:} A useful latent representation is one that describes the data well. Spatial (segmentation) maps can be considered a form of latent variable that allows visual inspection of what a network learns. At the same time, an easy (unsupervised) way to see whether a latent representation captures the data is to use a decoder to reconstruct the input.  In fact, even CycleGANs are autoencoders: they encode (and decode) the input via an intermediate output and thus inspire the design of our approach. Yet they have problems particularly when the intermediate output is discretised (a binary mask) and supervised losses are introduced. 
Their performance heavily depends on the weighting of the losses, as shown in Fig. \ref{fig:cyclegan_example}. If the segmentation loss is weighted higher than the reconstruction loss, it is not possible to reconstruct the input since the binary mask does not contain enough information for the transformation. When differently weighted, information is stored in the binary mask ruining semantics. This confirms findings of others, that a CycleGAN resolves the many-to-one/one-to-many problem by storing low-frequency information in the output image \cite{chu2017cyclegan}. 
We can see that the two losses are antagonistic, and a standard CycleGAN is not suitable as is.  We need to introduce variables that break the many-to-one problem, encouraging a balance between the losses to achieve good segmentation and reconstruction.
\begin{figure*}[t]
\centering
\includegraphics[width=\textwidth]{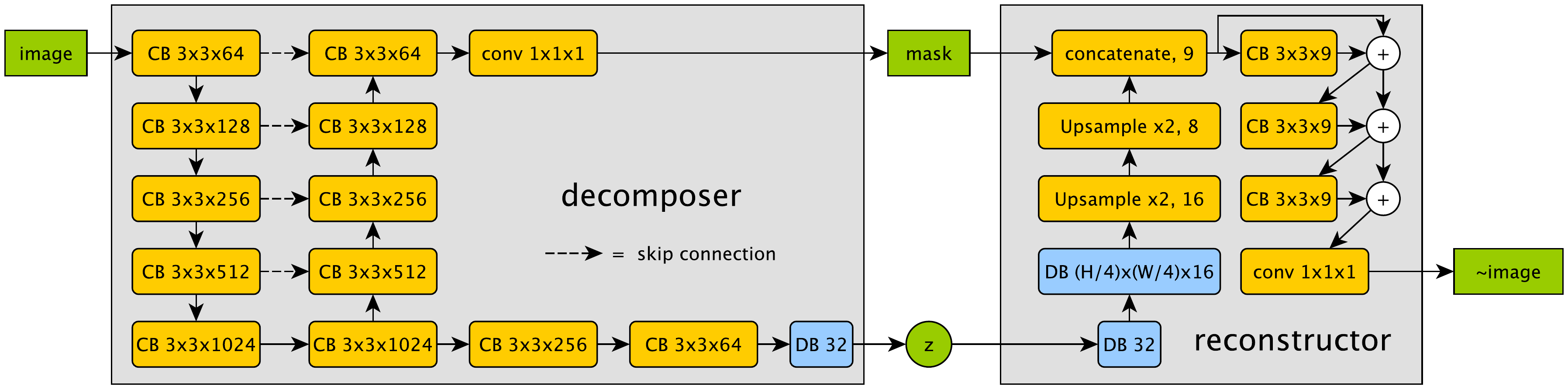}
\caption{Schematic of SDNet: an image is decomposed as a spatial representation of anatomy (in our case myocardial mask $M$) and a latent vector $Z$ that captures other anatomical and imaging characteristics. Both mask and $Z$ are used to reconstruct the input. The model consists of several convolutional (CB) and dense blocks (DB). BatchNormalization and LeakyRelu activations are used throughout.
\label{fig:schematic}}
\end{figure*}

\noindent \textbf{SDNet:} Our model is comprised of two interconnected neural networks, a ``decomposer'' and a ``reconstructor'', as illustrated in Fig.  \ref{fig:schematic}. The former decomposes an input 2D image (slice in a cine acquisition)  into two components: a spatial representation of the myocardium in the form of a binary mask, and a latent representation of the remaining anatomical and imaging features in the form of a vector. Thus, the mask is an image having pixel to pixel correspondences with the input and is inherently spatial, whereas the other representation is a vector representing information in a high level way that is not directly spatial. The reconstructor receives the two representations and aims to synthesise the original input image. Given a successful decomposition, the binary mask acts as a guide defining where the reconstructed myocardium should be. The role of the latent feature variable is then to learn some topology around the myocardium and fill the necessary intensity patterns, and allow for many-to-many mappings. 

\noindent \textbf{Costs:} More formally, let $f$ and $g$ be the decomposer and reconstructor. Given an image slice $X_i$, we aim to learn weights of $f$ to decompose into a mask $M$ and a 16 dimensional vector $Z$, that is $f(X_i) = \{f_M(X_i), f_Z(X_i)\} = \{M, Z\}$, and the weights of $g$ to remap the decomposition back to an image $g(f_M(X_i), f_Z(X_i))$.  

In a semi-supervised setup data comes from a labelled set $S_L = \{X_i, M_i\}_{i\in[1,N]}$ and an unlabelled set $S_U = \{X_j\}_{j\in[1,M]}$ where usually $M > N$. We now define the following losses.
Firstly, a reconstruction loss from autoencoding an image, $L_{rec}(f, g) = \mathop{\mathbb{E}}_X[\|X - g(f(X))\|_1]$. Secondly, two supervised losses when having images with corresponding masks $M_X$, $L_{M}(f) = \mathop{\mathbb{E}}_{X} [ Dice(M_X, f_M(X))]$, and $L_{I}(f, g) = \mathop{\mathbb{E}}_X[\|X - g(M_X, f_Z(X)))\|_1]$.
Finally, an adversarial loss using an image discriminator $D_X$, as $A_{I}(f, g, D_X) = \mathop{\mathbb{E}}_X [ D_X(g(f(X)))^2 + (D_X(X)-1)^2 ]$. Networks $f$ and $g$ are trained to maximise this objective against an adversarial discriminator trained to minimise it. 
Similarly, we define an adversarial loss using a mask discriminator $D_M$ as $A_{M}(f) = \mathop{\mathbb{E}}_{X,M} [ D_M(f_M(X))^2 + (D_M(M)-1)^2 ]$. Both adversarial losses are based on \cite{mao2017effectiveness}. The overall cost function is defined as:
\[
    \lambda_1 L_{M}(f) + 
    \lambda_2 A_{M}(f, D_M) + 
    \lambda_3 L_{rec}(f, g) +
    \lambda_4 L_{I}(f, g) +
    \lambda_5 A_{I}(f, g, D_X)
\]
The loss for images from the unlabelled set does not contain the first and fourth terms. The $\lambda$ are experimentally set to 10, 10, 1, 10 and 1 respectively.

\noindent \textbf{Implementation details:} The decomposer follows a U-Net \cite{ronneberger2015u} architecture (see Fig. \ref{fig:schematic}), and its last layer outputs a segmentation mask of the myocardium via a sigmoid activation function. The model's deep spatial maps contain downsampled image information, which is used to derive the latent vector $Z$ through a series of convolutions and fully connected layers, with the final output being passed through a sigmoid so $Z$ is bounded. Following this, an architecture with three residual blocks is employed as the reconstructor (see Fig. \ref{fig:schematic}).

The spatial and continuous representations are not explicitly made independent, so during training the model could still store all information needed for reconstructing the input as low values in the spatial mask, since finding a mapping from a spatial representation to an image is easier than combining two sources of information, namely the mask and $Z$. 
To prevent this, we apply a step function (i.e. a threshold) at the spatial input of the reconstructor to binarise the mask in the forward pass.
We store the original values and bypass the step function during back-propagation, and apply the updates to the original non-binary mask. Note that the binarisation of the mask only takes place at the input of the reconstructor network and is not used by the discriminator.

\section{Experiments and Discussion}
\subsection{Data and Baselines} \label{sec:exp_data}
\textbf{ACDC:} We use data from the 2017 ACDC Challenge\footnote{https://www.creatis.insa-lyon.fr/Challenge/acdc/index.html} containing cine-MR images from patients with various disease. Images were acquired in 1.5T or 3T MR scanners, with resolution between 1.22 and 1.68 $mm^2/pixel$ and the number of phases varying between 28 to 40 images per patient. We resample all volumes to 1.37 $mm^2/pixel$ resolution and normalise in the range [-1, 1]. \\
\textbf{QMRI:} We also use cine-MR data acquired at Edinburgh Imaging Facility QMRI with a 3T scanner, of 28 healthy patients, each having a volume of 30 frames. The spatial resolution is 1.406 $mm^2/pixels$ with a slice thickness 6mm, matrix size $256 \times 216$ and field of view $360mm \times 303.75mm$.

\noindent \textbf{Baselines: } We use as a \emph{fully-supervised baseline} a standard U-Net network trained with a Dice loss, similar to most participants of the ACDC challenge. 
We also consider a \emph{semi-supervised baseline}, shorthanded as GAN below, by adding a GAN loss to the supervised loss to allow adversarial training~\cite{luc2016semantic}. 

\begin{figure*}[t]
\centering
\includegraphics[width=1\textwidth]{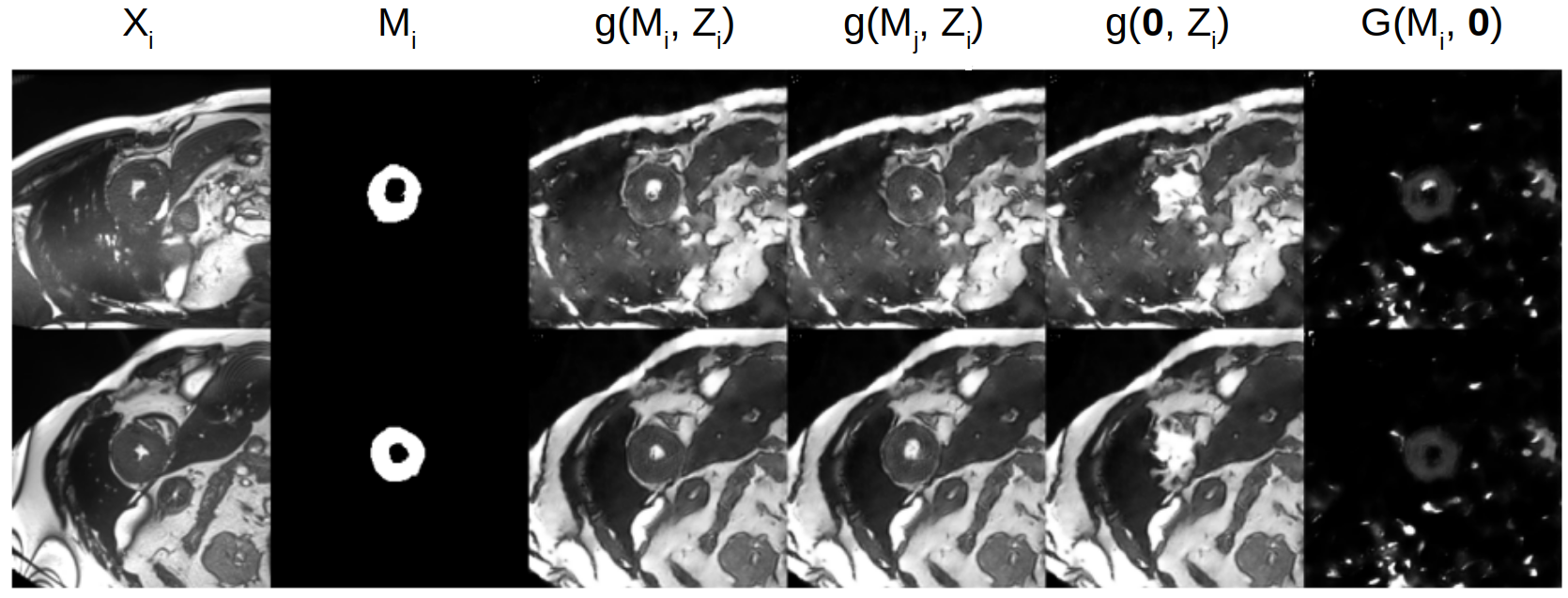}
\caption{Reconstructions using different $M_i$ and $Z_i$ combinations (see text for details).}
\label{fig:swap_masks}
\end{figure*}
\begin{figure*}[b]
\centering
\includegraphics[width=\textwidth]{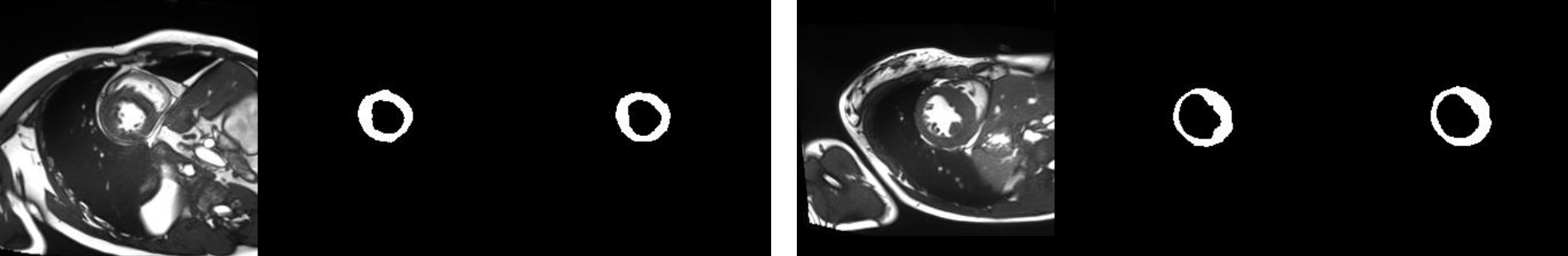}
\caption{Two examples of segmentation performance: input, prediction and ground truth.}
\label{fig:test_results}
\end{figure*}
\subsection{Latent space arithmetic}
\label{sec:exp_synthesis}
As a demonstration of our learned representation, in Fig. \ref{fig:swap_masks} we show reconstructions of input images from the training set using different combinations of masks and $Z$ components. In the first three columns, we show the original input with the predicted mask and the input's reconstruction. Next, we take the spatial representation $M_j$ from one image and combine it with the $Z_i$ component of the other image, and vice versa. As shown in the figure (4th column) the intensities and the anatomy around the myocardium remains unchanged, but the myocardial shape and position, which are encoded in the mask, change to that of the second image. The final two columns show reconstructions using a null mask (i.e. $M_i=\mathbf{0}$) and the correct $Z_i$ in 5th column, or using the original mask with a $Z_i=\mathbf{0}$ in 6th column. In the first case, the produced image does not contain myocardium, whereas in the second case the image contains only myocardium and no other anatomical or MR characteristics.

\subsection{Semi-Supervised Results} \label{sec:exp_semisupervised}
The utility of a factorised representation becomes evident in semi-supervised learning.  Qualitatively in Fig. \ref{fig:test_results} we can see that our method closely follows ground truth segmentation masks (example from ACDC held-out test set).  

To assess our performance quantitatively we train a variety of setups varying the number of labelled training images whilst keeping the unlabelled fixed (in both ACDC and QMRI cases). We train SDNet and the baselines (U-Net and GAN), test on held-out test sets, and use 3-fold cross validation (with 70\%, 15\%, 15\% of the volumes used in training, validation and test splits respectively). Results are shown in Table \ref{table:acdc}. For reference a U-Net trained with supervision on the full ACDC and QMRI datasets achieves a Dice score of 0.817 and 0.686 respectively. We can see that even when the number of labelled images is very low, our method is able to achieve segmentation accuracy considerably higher than the other two methods. As the number of labelled images increases, all models achieve similar accuracy.

\begin{table}
\centering
\begin{tabular}{l| ccccc | cccc}
\toprule
& \multicolumn{5}{c}{\textbf{ACDC}} & \multicolumn{4}{c}{\textbf{QMRI}} \\
Labelled images      & 284             & 142       & 68     &   34  & 11          & 157  & 78  & 39 & 19 \\
\midrule
\textbf{U-Net} & 0.782  & 0.657     & 0.581  &  0.356   & 0.026    & 0.686 & 0.681    & 0.441   & 0.368   \\
\textbf{GAN}   & \textbf{0.787}              & 0.727     & 0.648  & 0.365    & 0.080    &  \textbf{0.795}    & 0.756    & 0.580 &  0.061  \\
\textbf{SDNet} & 0.771  & \textbf{0.767}  & \textbf{0.731}  &  \textbf{0.678}   & \textbf{0.415}   &  0.794 & \textbf{0.772}  &  \textbf{0.686}   &  \textbf{0.424}   \\ 
\bottomrule
\end{tabular}
\caption{Myocardium Dice scores on ACDC and QMRI data. For training, 1200 unlabelled and varying numbers of labelled images were used. Masks for adversarial training came from the dataset, but do not correspond to any training images.}

\label{table:acdc}
\end{table}

\section{Conclusion}

We presented a method that decomposes images into spatial and (non-spatial) latent representations employing the cycle-consistency principle. To the best of our knowledge this is the first work to investigate spatial representation factorisation, in which one factor of the representation is inherently spatial, and thus well suited to spatial tasks. We  demonstrated its applicability in semi-supervised myocardial segmentation. In the low-data regime ($\approx 1\%$ of labelled with respect to unlabelled data) it achieves remarkable results, showing the power of the proposed learned representation. We leave as future work generative extensions, where we learn statistical distributions of our embeddings (as in VAEs). 
\\ \\
\noindent \textbf{Acknowledgements:}
This work was supported in part by the US National Institutes of Health (1R01HL136578-01) and UK EPSRC (EP/P022928/1). We also thank NVIDIA Corporation for donating a Titan X GPU. 

\printbibliography

\end{document}